\documentclass[lettersize,journal]{IEEEtran}
\usepackage{amsmath,amsfonts}
\usepackage{algorithmic}
\usepackage{algorithm}
\usepackage{array}
\usepackage[caption=false,font=normalsize,labelfont=sf,textfont=sf]{subfig}
\usepackage{textcomp}
\usepackage{stfloats}
\usepackage{url}
\usepackage{verbatim}
\usepackage{graphicx}
\usepackage{cite}
\usepackage{xcolor}
\allowdisplaybreaks[4]
\makeatletter
\renewcommand{\maketag@@@}[1]{\hbox{\m@th\normalsize\normalfont#1}}%
\makeatother
\begin{document}

\title{Distributed Leader Follower Formation Control of Mobile Robots based on Bioinspired Neural Dynamics and Adaptive Sliding Innovation Filter }


\markboth{}%
{Shell \MakeLowercase{\textit{et al.}}: A Sample Article Using IE
EEtran.cls for IEEE Journals}

\author{Zhe Xu,~\IEEEmembership{Member,~IEEE,} Tao Yan,~\IEEEmembership{Graduate Student Member,~IEEE,} \\Simon X. Yang,~\IEEEmembership{Senior Member,~IEEE,} S. Andrew Gadsden,~\IEEEmembership{Senior Member,~IEEE}
\thanks{This paper was accepted by IEEE Transactions on Industrial Informatics. \textbf{DOI: 10.1109/TII.2023.3272666}} 
\thanks{This work was supported by the Natural Sciences and Engineering Research Council (NSERC) of Canada. (\textit{Corresponding author: Simon X. Yang}) } 
\thanks{Zhe Xu, Tao Yan, and Simon X. Yang are with the Advanced Robotics and Intelligent Systems
(ARIS) Laboratory, School of Engineering, University of Guelph, Guelph, Ontario, Canada. e-mails: \{zxu02; tyan03; syang\}@uoguelph.ca}
\thanks{S. Andrew Gadsden is with the Intelligent and Cognitive Engineering (ICE) Lab, McMaster University, Hamilton, Ontario, Canada. e-mail: gadsden@mcmaster.ca}}
\maketitle

\begin{abstract}
This paper investigated the distributed leader follower formation control problem for multiple differentially driven mobile robots. A distributed estimator is first introduced and it only requires the state information from each follower itself and its neighbors. Then, we propose a bioinspired neural dynamic based backstepping and sliding mode control hybrid formation control method with proof of its stability. The proposed control strategy resolves the impractical speed jump issue that exists in the conventional backstepping design. Additionally, considering the system and measurement noises, the proposed control strategy not only removes the chattering issue existing in the conventional sliding mode control but also provides smooth control input with extra robustness. After that, an adaptive sliding innovation filter is integrated with the proposed control to provide accurate state estimates that are robust to modeling uncertainties. Finally, we performed multiple simulations to demonstrate the efficiency and effectiveness of the proposed formation control strategy. 
\end{abstract}


\section{Introduction}
\IEEEPARstart{T}{he} mobile robots have been widely used in many areas with the advancement of vehicular technologies. The cooperation among multiple mobile robots has vast applications such as target search, environment surveillance, and transportation \cite{Hu2022Multi-RobotSheet,Macwan2015ARescue}. Compared with single mobile robots, a multiple mobile robot system gives extra fault tolerance, extra flexibility, and completing tasks more efficiently\cite{Lin2021AdaptiveConstraints,Miao2022Low-ComplexityFeedback,Chang2022Fixed-TimePerformance}. Among all the applications of the multiple robot systems, formation control is one of the most fundamental and critical research areas. The objective of the robot’s formation control is to make each robot achieve the velocity required for the multiple robot system to maintain the prescribed formation pattern. 

There has been extensive research on the formation control of multiple mobile robots\cite{Han2022DistributedAttacks,Xu2014Behavior-basedRobots,Lee2018DecentralizedAvoidance,Moorthy2022DistributedApproach}; the commonly used coordination strategies include the virtual structure methods, the behavior based approaches, and the leader-following based approaches. It is observed that the leader-following approaches have attracted increasing attention due to their nice properties (e.g., easy to implement and analyze, good flexibility, etc.) \cite{Moorthy2022DistributedApproach,Miao2018DistributedRobots,Han2022DistributedAttacks,Hou2021DistributedLearning}. However, notice that in their works it is assumed that all followers can access the information of the leader's state, whereas most of the time accurate leader's state information could be challenging and complicated to obtain. Realistic factors such as communication bandwidth, the operating distance among the robots, and the total number of robots in the system can impact followers' ability to obtain the accurate leader's information. Therefore, it would be more practical to suppose only that a subset of mobile robots in the team can directly access the leader's state. To this end, it is necessary to investigate distributed estimation strategies for each robot in order to obtain the leaders' information. Several results can be found. Yu \textit{et al.} \cite{Yu2016DistributedConstraints} proposed a distributed formation control strategy for vehicles of the unicycle type subject to velocity constraints. After that, there is another research group proposed a distributed estimator to estimate the leader's state for each follower\cite{Miao2018DistributedRobots}. Another paper proposed a fixed time leader following consensus for multiple mobile robots \cite{Ning2020Fixed-TimeRobots}, in which the followers estimate the leader's state in a fixed time. Another group designed a distributed estimator to estimate the leader's state, which was later used to design the distributed kinematic controller that resolves the speed jump issue \cite{Moorthy2022DistributedApproach}. 

In addition to the leader's state estimation, the formation tracking control of multiple mobile robots is also a critical dimension that needs to be considered, especially under system and measurement noises. Roughly speaking, the formation tracking control strategies could be classified into the following five categories: 1) feedback linearization \cite{Yousuf2018FormationObservers}, 2) fuzzy logic and neural network control \cite{Medjoubi2021DesignApproach,Xiong2020Fixed-timeApproach}, 3) Optimal and model predictive control \cite{Rosenfelder2022CooperativeRobots,Worthmann2016ModelCosts}, 4) backstepping control \cite{Moorthy2022DistributedApproach,Wu2015MixedRobots} and 5) sliding mode control \cite{Zheng2017Leader-followerMode,Zhao2019LyapunovStrategy}. The feedback linearization method is easy to design and has been well developed. However, when initial tracking errors occur, this approach suffers from the speed jump problem that is impractical \cite{Besseghieur2020FromControl}. Besides, this method also produces large discontinuities in control commands if there is noise that occur. Neural networks and fuzzy logic control can deal with the speed jump issue and provide smooth control input. Nevertheless, these approaches require online learning or human knowledge, which are computationally expensive to implement \cite{Xiao2023IntegratedStrategy, Medjoubi2021DesignApproach}. Model predictive control is one of the popular approaches, which is able to optimize the online performance index and handle constraints. Although the predictive control methods can be used to theoretically achieve the expected performance, its results may not be guaranteed in practical cases \cite{Rosenfelder2022CooperativeRobots}. The main reason is that the ubiquitous disturbances and noise in real life will affect the results, but they were neglected in most of the literature in order to reduce computational efforts. Backstepping control is relatively easy to design and has been widely used in multiple mobile robot control \cite{Han2022DistributedAttacks}; however, this method also suffers from speed jump issues when initial tracking errors occur. Yang \textit{et al}.\cite{Yang2012ARobots} proposed a bioinspired neural dynamics based backstepping method that handles speed jump problems, making this method more practical in real-world applications. Following Yang's work \cite{Yang2012ARobots}, another group developed a formation tracking control algorithm for multiple mobile robots using bioinspired backstepping control to prevent speed jumps \cite{Moorthy2022DistributedApproach}. The sliding mode control methods are robust to disturbances, but knowingly suffer from 
 a chattering issue \cite{Zhao2019LyapunovStrategy}. In particular, the chattering and discontinuity phenomena in the control process get even worse under noises, thus making these methods undesirable in real-world applications.

In view of the noises, it becomes vital to have an efficient filter to provide accurate state estimates. Kalman filter (KF) and its variations have been undoubtedly regarded as the most effective techniques to estimate the states of the robots  \cite{Kim2016CubatureMulti-UAVs,Liu2021FormationApproach}. However, while these filters are proposed to give accurate state estimates under noises, they still lack the robustness when the modeling errors and uncertainties are present. Recently, an adaptive sliding innovation filter (ASIF) \cite{Newton2020Air-LUSI:Measurements} was newly developed in order to increase the accuracy yet maintain robustness with respect to modeling errors.

While a good amount of research efforts has been made in order to obtain practical and efficient distributed solutions for the multiple mobile robots formation tracking, there still are some aspects not properly addressed. For example, researchers in \cite{Moorthy2022DistributedApproach}, although employed a distributed state estimator in their design, only dealt with formation control at the kinematic level without consideration of disturbances and noises, which results in a restrictive design. As for the dynamic level, most of the conventional control methods, as mentioned above, suffer from drawbacks, including impractical speed jump, chattering, and sensitivity to noise \cite{Besseghieur2020FromControl,Han2022DistributedAttacks,Zhao2019LyapunovStrategy}. In view of these technical challenges, this paper presents a novel distributed formation tracking control approach for multiple mobile robots so that the aforementioned issues are addressed significantly. The main contributions of this research are summarized as follows:   


1. The distributed formation control problem of differentially driven mobile robots subject to communication limitations, disturbances, and noises is studied, in which only a subset of followers has the leader's state information. A novel distributed leader-following formation control strategy is developed aided with bioinspired neural dynamics.

2. Considering the communication limitations, a distributed estimator is developed for each robot to estimate the leader's posture information, and the leader's velocities are then estimated by the sliding mode approach without requiring the leader's acceleration information.

3. A novel hybrid bioinspired backstepping and sliding mode formation control strategy is proposed to deal with the speed jump issue in the conventional backstepping design and generate smooth torque control inputs free from chattering while robustness is maintained. More importantly, the bioinspired sliding mode control approach can provide smooth control activities even under the effects of the system and measurement noises. The overall stability is also proved.

4. The adaptive sliding innovation filter is integrated with the proposed distributed hybrid formation control method, aiming to provide accurate state estimates for each mobile robot even when modeling errors occur.

This paper is organized as follows; Section II provides preliminaries and formulates problems. Then, Section III designs the formation control method aided by the bioinspired neural dynamics. The ASIF is integrated with the proposed approach to provide accurate state estimates. After that, Section IV shows multiple simulation comparative results that to show the advantages of the proposed formation control strategy. Finally, Section V gives the conclusion and possible future works.

\section{Preliminaries and problem statement}
\subsection{Algebraic Graph Theory}
The graph theory is capable of modeling the communications among mobile robots. Consider a network that consists of $n$ followers; the digraph is defined as $\mathcal{G} = \{ V,E\} $ with a node set $V=\{1,2,...,n\}$ and an edge set $E \subseteq V \times V$. A directed edge $(i,j) \in E$ refers to the access from node $i$ to node $j$. Then, the adjacency matrix is defined $A ={[{a_{ij}}]_{n \times n}}$. If $A$ is a symmetric matrix, then the graph is undirected. The elements in $A$ are defined as follows ${a_{ij}}=1$ if $i,j\in E$ and  ${a_{ij}}=0$ otherwise. Furthermore, $a_{ii}$ is assumed to be $0$. The Laplacian matrix $L = {[{l_{ij}}]_{n \times n}}$, which is incorporated with $A$ is defined as\par
\vspace{-2ex}\begin{small}\begin{equation}
    {l_{ij}} = \left\{ {\begin{array}{*{20}{c}}
{ - {a_{ij}}}\qquad \qquad i \ne j\\
{\sum\limits_{i = 1,i \ne j}^n {{a_{ij}}} }\qquad i = j
\end{array}} \right.
\end{equation}\end{small}\normalsize
In this study, the leader mobile robot is labeled as $0$ and the followers are indexed by $1$ to $n$. Define the communications between the followers and the leader as $a = \left[ {\begin{array}{*{20}{c}}
{{a_{10}}}&{{a_{20}}}&{...}&{{a_{n0}}}
\end{array}} \right]^T$. If there is a communication connection between the leader and follower $j$ then $a_{j0}=1$, else $a_{j0}=0$. Then, we define $H \in {R^{n \times n}}$ as
\begin{equation}
    H = L + {\rm{diag}}(a). \label{2}
\end{equation}
In this paper, we the introduce following assumption and lemma based on previous studies \cite{Moorthy2022DistributedApproach,Miao2018DistributedRobots}.

\noindent{\bf Assumption 1.}
The undirected graph, denoted as $G=(V,E,A)$, illustrates the communications among $n$ followers. $G$ is connected if there is at least one element in $a$ is $1$, i.e., $a_{j0} = 1$.

\noindent{\bf Lemma 1.}
The matrix $H$ is only positive definite if the undirected graph $G$ is connected and at least one follower is the neighbor of the leader.

\subsection{Problem Formulation}
Consider a group of differentially driven mobile robots, the kinematics is described as
\begin{equation}
{{\dot x}_{i}} = {\upsilon _{i}}\cos {\theta _{i}} \quad
{{\dot y}_{i}} = {\upsilon _{i}}\sin {\theta _{i}} \quad
{{\dot \theta }_{i}} = {\omega _{i}}
\end{equation}
where ${x}_{i}$ and ${y}_{i}$ are coordinates and ${{\theta }_{i}}$ is the orientation of $i$-th robot with respect to the inertial frame; ${\upsilon _{i}}$ and ${\omega _{i}}$ are, respectively, the linear and angular velocity of $i$-th robot in body fixed frames. After the kinematics of the mobile robot is provided, the dynamics of the mobile robot is then given as\cite{Xu2020EnhancedFilter}\par
\vspace{-2ex}\begin{small}\begin{equation}
    \bar M(P_{i}){{\dot \xi }_{i}} = \bar B(P_{i})\tau_{i}  - \bar C(P_{i},\dot P_{i}){\xi _{i}}+\tau_{d,i}.
\end{equation}\end{small}\normalsize
where ${{\xi }_{i}}=[{\upsilon _{i}}, {\omega _{i}}]^T$; $\bar M(P_{i})$ is the inertial matrix; $\bar B(P_{i})$ is the transformation matrix; $C(P_{i},\dot P_{i})$ is the centrifugal and Coriolis matrix of the $i$-th mobile robot; ${\tau _{d,i}} = {\left[ {\begin{array}{*{20}{c}}
{{d _{1,i}}}&{{d _{2 ,i}}}
\end{array}} \right]^T}$ is the disturbances.\par

\noindent{\bf Assumption 2.}
The disturbances are assumed to be bounded such that
\begin{small}\begin{equation}
    \mathop {\max }\limits_{t \ge 0} \left| {{{d}_{1,i}}(t)} \right| \le {\psi _{1,i}}\quad \text{and}\quad \mathop {\max }\limits_{t \ge 0} \left| {{{d}_{2,i}}(t)} \right| \le {\psi _{2,i}}
\end{equation}\end{small}\normalsize
where $\psi _{1,i}$ and $\psi _{2,i}$ are positive constants. Then, the leader generates a reference signal that guides all the robots, the leader robot kinematics is given by
\begin{equation}
    {{\dot x}_{r}} = {\upsilon _{r}}\cos {\theta _{r}} \quad
{{\dot y}_{r}} = {\upsilon _{r}}\sin {\theta _{r}} \quad
{{\dot \theta }_{r}} = {\omega _{r}}.
\end{equation}
\noindent{\bf Assumption 3.}
Denote the virtual leader's linear and angular velocities as $\upsilon_{r}$ and $\omega_{r}$, and assume they are bounded. Additionally, assume that $\dot \upsilon_r$ and $\dot \omega_r$ are also bounded and there exist constants ${\iota _1}$ and ${\iota _2}$ such that
\begin{equation}
    \mathop {\max }\limits_{t \ge 0} \left| {{{\dot \upsilon }_r}(t)} \right| \le {\iota _1},\ \mathop {\max }\limits_{t \ge 0} \left| {{{\dot \omega }_r}(t)} \right| \le {\iota _2}.
\end{equation}
The leader follower formation control of mobile robot allows followers to follow the leader in a relative position, such that the distance between the leader and the $i$-th follower is described as $\left[ {\begin{array}{*{20}{c}}{\Delta {x_{i}}}&{\Delta {y_{i}}}\end{array}} \right]$, where ${\Delta {x_{i}}}$ and ${\Delta {y_{i}}}$ are respectively the desired distance from the $i$-th follower to the leader. In addition, the orientation of the followers shares a pattern similar to that of the leader.

In addition, the system and measurement noises have significant impacts on the control performance, and these noises are usually treated as Gaussian distributed with zero mean such that $(P_i({\alpha _{i,k}})\sim{\mathcal{N}}(0,{Q_{i,k}}))$, and $(P_i({\beta_{i,k}})\sim{\mathcal{N}}(0,{R_{i,k}}))$, where $\alpha _{i,k}$ and $\beta _{i,k}$ are the system and measurement noises at time step $k$, respectively. Parameters $Q_{i,k}$ and $R_{i,k}$ are respectively the system and measurement noise covariance, which are usually determined by the experiments and trials.

Therefore, the objective of this manuscript is to develop a distributed formation control strategy that achieves global bounded stability with smooth control inputs, such that
\begin{equation}
    \mathop {\lim }\limits_{t \to \infty } |{x_{i}} - {x_{r}} - \Delta {x_{i}}| \le \tau_i^x
\end{equation}
\begin{equation}
   \mathop {\lim }\limits_{t \to \infty } |{y_{i}} - {y_{r}} - \Delta {y_{i}}| \le \tau_i^y 
\end{equation}
\begin{equation}
    \mathop {\lim }\limits_{t \to \infty } |{\theta_{i}} - {\theta_{r}}| \le \tau_i^\theta 
\end{equation}
where $\tau _i^x$, $\tau _i^y $, and $\tau _i^\theta $ are arbitrary small thresholds.
\vspace{-8pt}
\section{Formation Control Design}
This section designs a novel leader and follower based formation control with the adaptive sliding innovation filter. The distributed estimator is first introduced. Then, based on the error dynamics, the bioinspired kinematic control is proposed using the backstepping approach to resolve impractical speed jump issues in the conventional design; after that, a sliding mode dynamic controller aided by the bioinspired neural dynamics is developed, which is free from chattering and provides smooth control input even under noises. The adaptive sliding innovation filter is then integrated with the proposed control strategy not only to provide accurate state estimate but also robust to faulty working conditions when modeling errors occur during state estimation processes.\vspace{-2ex}
\subsection{Distributed Estimator Design}
 The conventional design of the leader-follower formation control of multiple mobile robots makes the assumption that all followers have access to the leader's information. However, this conventional design faces difficulty when the communications among the followers are limited to local, therefore, not all followers are able to obtain the leader's state. Thus, this subsection develops a distributed estimator for every follower to estimate the leader's posture and velocities, which only requires access to neighborhood posture and velocities information.
 
 First, the posture estimation error for robot $i$ is defined as
 \begin{small}
 \begin{equation}
     {e_{ip}} = {a_{i0}}\left( {{P_{ir}} - {P_{r}}} \right)+ \sum\limits_{j = 1}^n {{a_{ij}}\left( {{P_{ir}} - {P_{jr}}} \right)}  \label{11}
 \end{equation} \end{small}\normalsize
 where $e_{ip}$ is the neighborhood errors of the $i$-th robot, ${P_{r}} = {\left[ {\begin{array}{*{20}{c}}{{x_{r}}}&{{y_{r}}}&{{\theta _{r}}}\end{array}} \right]^T}$ is the leader's posture; and ${P_{ir}} = {\left[ {\begin{array}{*{20}{c}}
{{x_{ir}}}&{{y_{ir}}}&{{\theta _{ir}}}
\end{array}} \right]^T}$ is the estimations of the leader posture from follower $i$ in the inertial frame. Then, the estimation strategy based on $e_{ip}$ is presented as\par
\begin{small} \begin{equation}
     {{\dot P}_{ir}} = \frac{1}{{{\xi _i}}}\left( { - {k_i}{e_{ip}} + \sum\limits_{j = 1}^n {{a_{ij}}{{\dot P}_{jr}} + {a_{i0}}{{\dot P}_{r}}} } \right)\label{12}
\end{equation}\end{small}\normalsize
where ${\xi _i} = \sum\nolimits_{j = 1}^n {{a_{ij}}}  + {a_{i0}}$, and $k_i$ is a $3\times3$ symmetric positive definite matrix. In Assumption 3, $\dot \upsilon_r$ is only assumed to be bounded, then the same approach that is given in \eqref{11} and \eqref{12} to estimate $\upsilon_r$ is not practical. Thus, to estimate the leader's velocities, a sliding mode approach has been developed. In \eqref{12}, the estimation of ${\dot P}_{ir}$ requires the information from ${{\dot P}_{jr}}$, which is obtained using the information from the velocity estimators that will be designed in the following part.
 
First, the linear velocity estimation error for robot $i$ is defined as
\vspace{-2ex}\begin{small}\begin{equation}
    {e_{i\upsilon }} =  {a_0}\left( {{\upsilon _{ir}} - {\upsilon _{r}}} \right)+\sum\limits_{j = 1}^n {{a_{ij}}\left( {{\upsilon _{ir}} - {\upsilon _{jr}}} \right)}  \label{13}
\end{equation}\end{small}\normalsize
where $\upsilon _{ir}$ and $\upsilon _{jr}$ are respectively the estimation of leader's linear velocity from follower $i$ and $j$. Based on the estimation error defined in \eqref{13}, the estimation law for ${\upsilon _{ir}}$ is then provided as\par
\vspace{-2ex}\begin{small}\begin{equation}
    {{\dot \upsilon }_{ir}} =  - {k_{a1}}{e_{i\upsilon }} - {k_{b1}}{\rm{sat}}\left( {{e_{i\upsilon }}} \right)\label{14}
\end{equation}\end{small}\normalsize
where positive constants ${k_{a1}}$ and ${k_{b1}}$ are both control parameters and ${k_{b1}} \ge \iota_1$. Using the same processes, define $\omega _{ir}$ and $\omega _{jr}$ are respectively the estimation of leader's angular velocity from follower $i$ and $j$, the estimation law for the angular velocity is given as\par
\vspace{-2ex}\begin{small}\begin{equation}
    {{\dot \omega }_{ir}} =  - {k_{a2}}{e_{i\omega }} - {k_{b2}}{\rm{sat}}\left( {{e_{i\omega }}} \right)\label{15}
\end{equation}\end{small}\normalsize
{\bf Theorem 1} If Assumption 1 holds, then ${P_{ir}}$, ${{\upsilon }_{ir}}$, and ${{\omega }_{ir}}$ respectively converge to ${P_{r}}$, ${{\upsilon }_{r}}$ and ${{\omega }_{r}}$ exponentially considering the estimation law \eqref{12}, \eqref{14}, and \eqref{15}.

{\textit{Proof}}
In order to prove Theorem 1, consider the following Lyapunov candidate function $\frac{1}{2}e_{ip}^T{e_{ip}}$. The derivative of the proposed Lyapunov candidate function is calculated as\par
\vspace{-2ex}\begin{small}\begin{equation}
\begin{aligned}
e_{ip}^T{{\dot e}_{ip}} &= e_{ip}^T\left( {\sum\limits_{j = 1}^n {{a_{ij}}\left( {{{\dot P}_{ir}} - {{\dot P}_{jr}}} \right)}  + {a_{i0}}\left( {{{\dot P}_{ir}} - {{\dot P}_{r}}} \right)} \right)\\
 &= e_{ip}^T\left( {{\xi _i}{{\dot P}_{ir}} - \sum\limits_{j = 1}^n {{a_{ij}}{{\dot P}_{jr}} - } {a_{i0}}{{\dot P}_{r}}} \right)=  - e_{ip}^T{k_i}e_{ip}
\end{aligned}
\end{equation}\end{small}\normalsize
Since $k_i$ is symmetric positive definite, one can conclude that $e_{ip}$ converges to zero. Additionally, define ${e_p} = \left[ {\begin{array}{*{20}{c}}
{e_{1p}^T}&{e_{2p}^T}&{...}&{e_{np}^T}\end{array}} \right]^T$ and ${e_q} = \left[ {\begin{array}{*{20}{c}}
{e_{1q}^T}&{e_{2q}^T}&{...}&{e_{nq}^T}
\end{array}} \right]^T$, where $e_{iq}$ is defined as ${e_{iq}} = {P_{ir}} - {P_{r}}$, then it is found that\par
\vspace{-2ex}\begin{small}\begin{equation}
    {e_p} = \left( {H \otimes {I_3}} \right){e_q}
\end{equation}\end{small}\normalsize
Based on \eqref{2}, $H$ is a symmetric positive definite matrix if Assumption 1 and Lemma 1 hold. Thus, as ${e_p}$ exponentially converges to zero, ${e_q}$ also converges to zero.\par

Define ${{\hat \upsilon }_i} = {\upsilon _{ir}} - {\upsilon _{r}}$, based on the estimation law defined in \eqref{14}, it is found that
\begin{equation}
    {{\dot {\hat \upsilon }_i}} =  - {k_{a1}}{e_{i\upsilon }} - {k_{b1}}{\rm{sat}}\left( {{e_{i\upsilon }}} \right) - {{\dot \upsilon }_{r}}\label{18}
\end{equation}
Then, define ${e_\upsilon } = {\left[ {\begin{array}{*{20}{c}}
{{e_{1\upsilon }}}&{{e_{2\upsilon }}}&{...}&{{e_{n\upsilon }}}
\end{array}} \right]^T}$ and $\hat \upsilon  = {\left[ {\begin{array}{*{20}{c}}
{{{\hat \upsilon }_1}}&{{{\hat \upsilon }_2}}&{...}&{{{\hat \upsilon }_n}}
\end{array}} \right]^T}$. It is obtained that\par
\vspace{-2ex}
\begin{equation}
    {e_\upsilon } = H\hat \upsilon \label{19}
\end{equation}
\normalsize
Then, based on \eqref{18} and \eqref{19}, it is easy to verify that
\begin{small}\begin{equation}
    {\dot {\hat \upsilon} }  =  - {k_{a1}}{e_\upsilon } - {k_{b1}}{\rm{sat}}\left( {{e_\upsilon }} \right) - {{\dot \upsilon }_{r}}{1_n}
\end{equation}\end{small}\normalsize
Similarly, define ${{\hat \omega }_i} = {\omega _{ir}} - {\omega _{r}}$. Based on \eqref{15}, the estimation law for the angular velocity can be rewritten as\par
\begin{small}\begin{equation}
    \dot {\hat \omega}  =  - {k_{a2}}{e_\omega } - {k_{b2}}{\rm{sat}}\left( {{e_\omega }} \right) - {{\dot \omega }_{r}}{1_n}
\end{equation}\end{small}\normalsize
where ${e_\omega } = {\left[ {\begin{array}{*{20}{c}}
{{e_{1\omega }}}&{{e_{2\omega }}}&{...}&{{e_{n\omega }}}
\end{array}} \right]^T}$ and $\hat \omega  = {\left[ {\begin{array}{*{20}{c}}
{{{\hat \omega }_1}}&{{{\hat \omega }_2}}&{...}&{{{\hat \omega }_n}}
\end{array}} \right]^T}$. Once again, it is easy to find that ${e_\omega } = H\hat \omega$.
In order to prove the stability, the Lyapunov candidate function is proposed as $V_1=\frac{1}{2}{{\hat \upsilon }^T}H\hat \upsilon$, and its time derivative is calculated as\par
\vspace{-2ex}\begin{small}\begin{equation}
    \begin{aligned}
  {{\dot V}_1} &= {{\hat \upsilon }^T}H\left( { - {k_{a1}}{e_\upsilon } - {k_{b1}}{\rm{sat}}\left( {{e_\upsilon }} \right) - {{\dot \upsilon }_{r}}{1_n}} \right)\\
 &=  - {k_{a1}}{{\hat \upsilon }^T}HH\hat \upsilon  - {k_{b1}}{\left( {H\hat \upsilon } \right)^T}{\rm{sat}}\left( {H\hat \upsilon } \right) - {{\hat \upsilon }^T}H{{\dot \upsilon }_{r}}{1_n}  
    \end{aligned}\label{23}
\end{equation}\end{small}If ${\rm{sat}}({{e_\upsilon }})$ is saturated, which means the elements in ${{H\hat \upsilon } }\ge 1$, then \eqref{23} satisfies
\begin{small}\begin{equation}
    {{\dot V}_1} \le  - {k_{a1}}{{\hat \upsilon }^T}HH\hat \upsilon  - \left( {{k_{b1}} - {\iota _1}} \right){\left\| {H\hat \upsilon } \right\|_1}
\end{equation}\end{small}where ${\left\| {H\hat \upsilon } \right\|_1}$ is the first norm of vector ${H\hat \upsilon }$. we use the fact that ${k_{b1}}\ge \iota _1$ and $H$ is a symmetric positive definite matrix to arrive at the conclusion that the estimation law for the linear velocity is asymptotically stable.

\noindent{\bf Lemma 2.} Based on Cauchy-Schwarz inequality, the following inequality holds.

\begin{small}\begin{equation}
    {\left\| {H\hat \upsilon } \right\|_1} \le \sqrt n \left\| {H\hat \upsilon } \right\|
\end{equation}\end{small}If the elements in ${H\hat \upsilon }< 1$, by applying Lemma 2, \eqref{23} is calculated as
\begin{small}\begin{equation}
   \begin{aligned}
     {{\dot V}_1} &\le  - \left( {{k_{a1}} + {k_{b1}}} \right){\left\| {H\hat \upsilon } \right\|^2} + {\iota _1}\sqrt n \left\| {H\hat \upsilon } \right\|\\
 &=  - \left( {{k_{a1}} + {k_{b1}} - {\gamma _1}} \right){\left\| {H\hat \upsilon } \right\|^2}\quad{\text{whenever}}\left\| {H\hat \upsilon } \right\| \ge {\mu _1}\label{26}
   \end{aligned}
\end{equation}\end{small}where ${\gamma _1}$ is an arbitrary number within the interval $(0,{k_{a1}} + {k_{b1}})$ and ${\mu _1} = \left({\iota _1}\sqrt n  \right)/{{\gamma _1}} $. Thus, the linear velocity estimation strategy is input-to-state stable with respect to input $||{\dot \upsilon }_{r}||$. In particular, if $t \to \infty$ and $\left\| {{{\dot \upsilon }_{r}}} \right\| \to 0$, the estimation error $\hat \upsilon$ exponentially converges to zero.

The Lyapunov candidate function for the angular estimation law is proposed as $V_2=\frac{1}{2}{{\hat \omega }^T}H\hat \omega$, by applying Assumption 1 and Lemma 2, following the same processes from \eqref{23} to \eqref{26}. It is found that if $\left|| {H\hat \upsilon } |\right|\ge1$, the angular velocity estimation strategy is asymptotically stable, if $\left|| {H\hat \upsilon } |\right|<1$, $\dot V_2$ yields\par
\noindent\begin{small}
\begin{equation}
    \begin{aligned}
{{\dot V}_2} &\le  - \left( {{k_{a2}} + {k_{b2}}} \right){\left\| {H\hat \upsilon } \right\|^2} + {\iota _2}\sqrt n \left\| {H\hat \upsilon } \right\|\\
 &=  - \left( {{k_{a2}} + {k_{b2}} - {\gamma _2}} \right){\left\| {H\hat \upsilon } \right\|^2}\quad{\text{whenever}}\left\| {H\hat \upsilon } \right\| \ge {\mu _2}
\end{aligned}
\end{equation}
\end{small}\normalsize where ${\gamma _2}$ is an arbitrary number within the interval $(0,{k_{a2}} + {k_{b2}})$ and ${\mu _2} = \left({\iota _2}\sqrt n  \right)/{{\gamma _2}} $. Therefore, the estimation law for the angular velocity is input-to-state stable with respect to ${\dot \omega }_{r}$. In addition, if $t \to \infty$ and $\left\| {{{\dot \omega }_{r}}} \right\| \to 0$, the estimation error $\hat \omega$ exponentially converges to zero asymptotically. The proof of Theorem 1 is finished.

\subsection{Bioinspired Backstepping Controller Design}
The distributed bioinspired backstepping kinematic controller is designed in this subsection to achieve formation control objectives using the estimated state of the leader. In addition, the leader's state information is only available to a subset of followers. The tracking error of follower $i$ in the inertial frame is defined as\par 
\begin{small}\begin{equation}
    \begin{array}{l}
{e_{ix}} = {x_{i}} - {x_{ir}} - \Delta {x_i}\\
{e_{iy}} = {y_{i}} - {y_{ir}} - \Delta {y_i}\\
{e_{i\theta }} = {\theta _{i}} - {\theta _{ir}}
\end{array}
\end{equation}
\end{small}\normalsize
where $\Delta {x_i}$ and $\Delta {y_i}$ are respectively the distances between the $i$-th follower and the leader in $x$ and $y$ directions. It is obvious that ${x_{ir}}$, ${y_{ir}}$, and ${\theta_{ir}}$ exponentially converge to ${x_{r}}$, ${y_{r}}$, and ${\theta_{r}}$, respectively.
Then, the tracking error between the body fixed frame and the inertial frame is calculated as \par
\noindent\begin{small}
\begin{equation}
    \left[ {\begin{array}{*{20}{c}}
{{{\hat x}_{i}}}\\
{{{\hat y}_{i}}}\\
{{{\hat \theta }_{i}}}
\end{array}} \right] = \left[ {\begin{array}{*{20}{c}}
{\cos {\theta _{i}}}&{\sin {\theta _{i}}}&0\\
{ - \sin {\theta _{i}}}&{\cos {\theta _{i}}}&0\\
0&0&1
\end{array}} \right]\left[ {\begin{array}{*{20}{c}}
{{e_{ix}}}\\
{{e_{iy}}}\\
{{e_{i\theta }}}
\end{array}} \right]\label{29}
\end{equation}
\end{small}\normalsize
where ${{\hat x}_{i}}$, ${{\hat y}_{i}}$, and ${{\hat \theta}_{i}}$ are respectively the tracking errors in the driving, lateral directions, and orientation. By taking the derivative of \eqref{29}, we have\par
\noindent\begin{small}\begin{equation}
\left[ {\begin{array}{*{20}{c}}
{{{\dot {\hat x}}_{i}}}\\
{{{\dot {\hat y}}_{i}}}\\
{{{\dot {\hat \theta}}_{i}}}
\end{array}} \right] = \left[ {\begin{array}{*{20}{c}}
{{\omega _{i}}{{\hat y}_{i}} - {\upsilon _{i}} + {\upsilon _{ir}}\cos {{\hat \theta }_{i}} + {\Omega _{ix}}}\\
{ - {\omega _{i}}{{\hat x}_{i}} + {\upsilon _{ir}}\sin {{\hat \theta }_{i}}  + {\Omega _{iy}}}\\
{{\omega _{i}} - {\omega _{ir}}}\label{30}
\end{array}} \right]
\end{equation}
\end{small}where ${\Omega _{ix}}$ and ${\Omega _{iy}}$ are respectively defined as
\begin{small}
\begin{equation}
    \begin{aligned}
{\Omega _{ix}} &= ({\upsilon _{ir}}\sin {\theta _{ir}} - {{\dot y}_{ir}})\sin {\theta _{i}} + ( {{\upsilon _{ir}}\cos {\theta _{ir}}} {- {{\dot x}_{ir}}} )\cos {\theta _{i}}\\
{\Omega _{iy}} &= ({\upsilon _{ir}}\sin {\theta _{ir}} - {{\dot y}_{ir}})\cos {\theta _{i}} - ( {{\upsilon _{ir}}\cos {\theta _{ir}} }{- {{\dot x}_{ir}}} )\sin {\theta _{i}}.\label{31}
    \end{aligned}
\end{equation}
\end{small}\normalsize

The conventional backstepping control suffers from speed jump issue, In order to resolve this problem the bioinspired neural dynamics is integrated with backstepping control to avoid the velocity jump issue. The bioinspired neural dynamics was initially proposed in the 1950s \cite{Hodgkin1952ANerve} for a nervous system in a membrane model, which was further developed by Grossberg \cite{Grossberg1988NonlinearArchitectures} to describe an adaptive dynamic behavior of individuals. Researchers have applied this model in different fields, which include robotics \cite{Yang2012ARobots,Moorthy2022DistributedApproach}.

The shunting model that is originally derived from Grossberg's neural dynamics model is provided as
\begin{equation}
    {\dot V_{si}} =  - {A_i}{V_{si}} + \left( {{B_i} - {V_{si}}} \right)f\left( {{{\hat x}_{i}}} \right) - \left( {{D_i} + {V_{s1}}} \right)g\left( {{{\hat x}_{i}}} \right)\label{34}
\end{equation}
where $f\left( {{{\hat x}_{i}}} \right) = \max \left( {{{\hat x}_{i}},0} \right)$ and $g\left( {{{\hat x}_{i}}} \right) = \max \left( {0, - {{\hat x}_{i}}} \right)$; $V_{s1}$ is the output of the shunting model; $A_i$ is the passive decay rate; $B_i$ and $D_i$ are respectively the upper and lower bounds of the output. 
We use this shunting model to replace the error term ${{\hat x}_{i}}$. Then, the bioinspired backstepping control is proposed as\par
\begin{small}
\begin{equation}
   {\upsilon _{ci}} = {C_{1}}{V_{si}} + {\upsilon _{ir}}\cos {{\hat \theta }_{i}} \label{35}
\end{equation}
\begin{equation}
   {\omega _{ci}} = {\omega _{ir}} + {C_{2}}{\upsilon _{ir}}{{\hat y}_{i}} + {C_{3}}{\upsilon _{ir}}\sin {{\hat \theta }_{i}}\label{36}
\end{equation}
\end{small}\normalsize
where $C_1$, $C_2$, and $C_3$ are positive design parameters.

\noindent{\bf Remark 1.} The initial speed jump issue in the conventional backstepping control is caused by the error term ${C_1}{{\hat x}_{i}}$. If there exists an initial tracking error, the velocity demand generated from the conventional backstepping control will not start from zero, which implies the initial torque demand tends to be infinitely large and impractical. The larger the control parameter ${C_1}$, the larger the initial speed jump occurs. In contrast, the shunting model acts like a low pass filter, which not only yields the output to start from zero but also bounds the output between a finite interval $(-D_i, B_i)$, making the maximum velocity command as $\max \left( {{\upsilon _{c,i}}} \right) = {C_1}{B_i} + {\upsilon _{ir}}$. Thus, a well designed bioinspired backstepping control could avoid speed jump issue and restrict the maximum velocity.
\subsection{Bioinspired Sliding Mode Control Design}
Followed by the design of bioinspired backstepping control. This subsection developed a bioinspired sliding mode control that uses the same neural dynamics. Define the error between the velocity commands and the estimated velocities that are generated from the filter as\par
\begin{small}
\begin{equation}
    \left[ {\begin{array}{*{20}{c}}
{{e_{\eta 1,i}}}\\
{{e_{\eta 2,i}}}
\end{array}} \right] = \left[ {\begin{array}{*{20}{c}}
{{\upsilon _{ci}} - { \upsilon _{i}}}\\
{{\omega _{ci}} - { \omega _{i}}}
\end{array}} \right]
\end{equation}
\end{small}\normalsize
where ${e_{\eta 1,i}}$ and ${e_{\eta 2,i}}$ are respectively defined as the linear and angular velocity errors. It is noted that the estimated velocity obtained from the filter can be converged to its actual value in some mean square sense. $ {\upsilon _{i}}$ and ${\omega _{i}}$ are respectively the actual linear and angular velocities. By setting the origin of the body fixed frame and gravity center to be at the same position, the conventional design of the sliding mode dynamic controller is given as\par
\noindent\begin{small}
\begin{equation}
\begin{aligned}
 \hspace{-3pt}{\tau _{L,i}} \!=\! \frac{{{m_i}{r_i}}}{2}\left( {{{\dot \upsilon }_{ci}} \!+\! {C_{a,i}}\rm{sign}{(e_{\eta 1,i})}} \right) \!-\! \frac{{{I_i}{r_i}}}{{2{c_i}}}({{\dot \omega }_{ci}} \!+\! {C_{b,i}}\rm{sign}{(e_{\eta 2,i})})
\end{aligned}
\end{equation}
\begin{equation}
\begin{aligned}
  \hspace{-3pt}{\tau _{R,i}} \!=\! \frac{{{m_i}{r_i}}}{2}\left( {{{\dot \upsilon }_{ci}} \!+\! {C_{a,i}}\rm{sign}{(e_{\eta 1,i})}} \right) \!+\! \frac{{{I_i}{r_i}}}{{2{c_i}}}({{\dot \omega }_{ci}} \!+\! {C_{b,i}}\rm{sign}{(e_{\eta 2,i})})
\end{aligned}
\end{equation}
\end{small}\normalsize 
where ${C_{a,i}}$ and ${C_{b,i}}$ are the positive control parameters.
The above conventional sliding mode approach suffers from the chattering issue, which generates discontinuous torque commands that are impossible to reach by actuators. Thus, the chattering term is replaced by the shunting model, defining the shunting models for linear and angular velocities as\par
\vspace{-1ex}\noindent\begin{small}
\begin{equation}
    {\dot V_{s1,i}}\! =\!  - {A_{1,i}} + \!\left( {{B_{1,i}} \!- {V_{s1,i}}} \right)f\left( {{e_{\eta 1,i}}} \right) - \left( {{D_{1,i}} \!+\! {V_{s1,i}}} \right)g\left( {{e_{\eta 1,i}}} \right)\label{40}
\end{equation}
\begin{equation}
    {\dot V_{s2,i}}\! =\!  - {A_{2,i}}+\! \left( {{B_{2,i}} \! - {V_{s2,i}}} \right)f\left( {{e_{\eta 2,i}}} \right) - \left( {{D_{2,i}} \!+\! {V_{s2,i}}} \right)g\left( {{e_{\eta 2,i}}} \right)\label{41}
\end{equation}
\end{small}\normalsize 
Then, the bioinspired sliding mode torque control is proposed as
\vspace{-1ex}\noindent\begin{small}
\begin{equation}
    {\tau _{L,i}} = \frac{{{m_i}{r_i}}}{2}\left( {{{\dot \upsilon }_{ci}} + {C_{a,i}}{V_{s1i}}} \right) - \frac{{{I_i}{r_i}}}{{2{c_i}}}({{\dot \omega }_{ci}} + {C_{b,i}}{V_{s2i}})\label{42}
\end{equation}
\begin{equation}
    {\tau _{R,i}} = \frac{{{m_i}{r_i}}}{2}\left( {{{\dot \upsilon }_{ci}} + {C_{a,i}}{V_{s1i}}} \right) + \frac{{{I_i}{r_i}}}{{2{c_i}}}({{\dot \omega }_{ci}} + {C_{b,i}}{V_{s2i}})\label{43}
\end{equation}
\end{small}\noindent{\bf Remark 2.} The chattering issue in the conventional sliding control has been eliminated with the application of the shunting model. Although there are other controllers such as saturation functions to avoid chattering issue, the control parameters ${C_{a,i}}$ and ${C_{b,i}}$ in the conventional design does not completely resolve the discontinuities in torque control command when considering noises. The shunting model in this case acts like a low pass filter, which yields smooth control commands.

\noindent{\bf Remark 3.} In order to overcome the disturbances, the control parameters ${C_{a,i}}$, ${C_{b,i}}$ tend to be large, which could potentially provide explosive torque command. However, the shunting model that is applied to sliding mode control offers extra robustness to disturbances without the requirements of large ${C_{a,i}}$ and ${C_{b,i}}$. In addition, the shunting models are also bounded between $\left( { - {D_{1i}},{B_{1i}}} \right)$ and $\left( { - {D_{2,i}},{B_{2,i}}} \right)$ for the control of linear and angular velocities, respectively.
\subsection{Stability Analysis}
This subsection proves the stability of the proposed formation control strategy, the Lyapunov candidate function for the bioinspired backstepping control is given as\par
\begin{small}
\begin{equation}
    {V_3} = \sum\limits_{i = 1}^n {\left( {\frac{1}{2}\hat x_{i}^2 + \frac{1}{2}\hat y_{i}^2 + \frac{1}{{{C_{2}}}}\left( {1 - \cos {{\hat \theta }_{i}}} \right) + \frac{{{C_{1}}}}{{2{B_i}}}}V_{si}^2 \right)} \label{44}
\end{equation}
\end{small}\normalsize
Based on \eqref{31}, \eqref{35}, and \eqref{36}, and setting $B_i=D_i$, the time derivative of \eqref{44} is calculated as\par
\noindent\begin{small}
\begin{equation}
\begin{aligned}
     {{\dot V}_3} &= \sum\limits_{i = 1}^n \left({{\dot {\hat x}}_{i}}{{\hat x}_{i}} + {{\dot {\hat y}}_{i}}{{\hat y}_{i}} + \frac{1}{{{C_{2}}}}{{\dot {\hat \theta} }_{i}}\sin {{\hat \theta }_{i}} + \frac{{{C_{1}}}}{{{B_i}}}{{\dot V}_{si}}{V_{si}}\right) \\
     &  = \sum\limits_{i = 1}^n {{C_{1}}\left( { - {{\hat x}_{i}} + f\left( {{{\hat x}_{i}}} \right) - g\left( {{{\hat x}_{i}}} \right)} \right)} {V_{si}} + \sum\limits_{i = 1}^n {\frac{{{C_{1}}}}{{{B_i}}}} ( \! - \!{A_i} - f\left( {{{\hat x}_{i}}} \right)\\&\quad - g\left( {{{\hat x}_{i}}} \right) )V_{si}^2  - \sum\limits_{i = 1}^n {\frac{{{C_3}}}{{{C_2}}}{\upsilon _{ir}}{{\sin }^2}{{\hat \theta }_i}}+ \sum\limits_{i = 1}^n {\left( {{{\hat x}_i}{\Omega _{ix}} + {{\hat y}_i}{\Omega _{iy}}} \right)} \label{45}
\end{aligned}   
\end{equation}
\end{small}\normalsize
If $ {{{\hat x}_{i}}}\ge0$, then $f\left( {{{\hat x}_{i}}} \right)={{{\hat x}_{i}}}$ and $g\left( {{{\hat x}_{i}}} \right)=0$. Thus, \par
\noindent\begin{equation}
 { { - {{\hat x}_{i}} + f\left( {{{\hat x}_{i}}} \right) - g\left( {{{\hat x}_{i}}} \right)} }=0 \label{46}   \vspace{-4pt}
\end{equation}
If $ {{{\hat x}_{i}}}<0$, then $g\left( {{{\hat x}_{i}}} \right)={{-{\hat x}_{i}}}$ and $f\left( {{{\hat x}_{i}}} \right)=0$, \eqref{46} also equals zero. In addition, it is easy to verify that $ - {A_i} - f\left( {{{\hat x}_{i}}} \right) - g\left( {{{\hat x}_{i}}} \right)\le0$. 

It should be noted that both ${\Omega _{ix}}$ and ${\Omega _{iy}}$ exponentially converge to zero, thus $\dot V_3\le0$. Furthermore, based on \eqref{34}, ${{{V_{si}}}}\to 0$ as $t\to \infty$, then ${{\hat x}_{i}}$ approaches zero as well. From term ${\frac{{{C_3}}}{{{C_2}}}{\upsilon _{ir}}{{\sin }^2}{{\hat \theta }_i}}$ in \eqref{45}, ${{\hat \theta }_i}\to 0$ as time$\to\infty$, then, based on \eqref{36}, ${C_2}{\upsilon _{ir}}{{\hat y}_{i}}$, ${{\hat y}_{i}}\to0$ as well. Therefore, the proposed control approach is asymptotically stable.  
 As for the bioinspired sliding mode controller, the Lyapunov candidate function is proposed as\par
\vspace{-2ex}\begin{small}
\begin{equation}
    {V_4} = \frac{1}{2}\sum\limits_{i = 1}^n {\left( {e_{\eta 1,i}^2 + e_{\eta 2,i}^2 + {C_{a,i}}V_{s1,i}^2 + {C_{bi}}V_{s2,i}^2} \right)}\label{50}
\end{equation}
\end{small}\normalsize
Then, the derivative of \eqref{50} is written as
\begin{small}
\begin{equation}
\begin{aligned}
{{\dot V}_4} \!= \sum\limits_{i = 1}^n {\left( {{{\dot e}_{\eta 1,i}}{e_{\eta 1,i}} \!+\! {{\dot e}_{\eta 2,i}}{e_{\eta 2,i}}\! +\! {C_{a,i}}{{\dot V}_{s1,i}}{V_{s1,i}} \!+\! {C_{bi}}{{\dot V}_{s2,i}}{V_{s2,i}}} \right)} 
 \label{48}  
\end{aligned}
\end{equation}
\end{small}\normalsize
By substituting \eqref{42} and \eqref{43} into \eqref{48}, it becomes\par
\vspace{-3ex}\begin{small}
    \begin{align}
       \hspace{-3pt} {{\dot V}_4}&=
         \sum\limits_{i = 1}^n {\frac{{{C_{a,i}}}}{{{B_{1,i}}}}\left( { - {B_{1,i}}{e_{\eta 1,i}} + {B_{1,i}}f\left( {{e_{\eta 1,i}}} \right) - {D_{1,i}}g\left( {{e_{\eta 1,i}}} \right)} \right)} {V_{s1,i}}\notag\\ 
         &+ \sum\limits_{i = 1}^n {\frac{{{C_{a,i}}}}{{{B_{1,i}}}}} \left( { - {A_{1,i}} - f\left( {{e_{\eta 1,i}}} \right) - g\left( {{e_{\eta 1,i}}} \right)} \right)V_{s1,i}^2+\sum\limits_{i = 1}^n {{e_{\eta 1,i}}{d_{1,i}}} \notag\\   &+\sum\limits_{i = 1}^n {\frac{{{C_{b,i}}}}{{{B_{2,i}}}}\left( { - {B_{2,i}}{e_{\eta 2,i}} + {B_{2,i}}f\left( {{e_{\eta 2,i}}} \right) - {D_{2,i}}g\left( {{e_{\eta 2,i}}} \right)} \right)} {V_{s2,i}}\notag\\
         & + \sum\limits_{i = 1}^n {\frac{{{C_{b,i}}}}{{{B_{2i}}}}} \left( { - {A_{2,i}} - f\left( {{e_{\eta 2,i}}} \right) - g\left( {{e_{\eta 2,i}}} \right)} \right)V_{s2,i}^2+\sum\limits_{i = 1}^n {{e_{\eta 2,i}}{d_{2,i}}}\label{49}
    \end{align}
\end{small}\normalsize
Once again, using the characteristic of the shunting model, by assuming $B_{1i}=D_{1i}$ and $B_{2i}=D_{2i}$, it is calculated that\par
\begin{small}

\begin{equation}
\begin{aligned}
      - {B_{1i}}{e_{\eta 1,i}} + {B_{1,i}}f\left( {{e_{\eta 1,i}}} \right) - {D_{1,i}}g\left( {{e_{\eta 1,i}}} \right) = 0\\
     - {B_{2i}}{e_{\eta 2,i}} + {B_{2,i}}f\left( {{e_{\eta 2,i}}} \right) - {D_{2,i}}g\left( {{e_{\eta 2,i}}} \right) = 0  
\end{aligned}
\end{equation}
\end{small}\normalsize
Then, \eqref{49} is rewritten as
\begin{small}\begin{equation}
\begin{aligned}
      {{\dot V}_4}&=\sum\limits_{i = 1}^n {\left( {\left( { - \frac{{{A_{1,i}}{C_{a,i}}}}{{{B_{1,i}}}} - \frac{{{C_{a,i}}\left| {{e_{\eta 1,i}}} \right|}}{{{B_{1,i}}}}} \right)V_{s1,i}^2 + {e_{\eta 1,i}}{d_{1,i}}} \right)}  \\ &+\sum\limits_{i = 1}^n {\left( {\left( { - \frac{{{A_{2,i}}{C_{b,i}}}}{{{B_{2,i}}}} - \frac{{{C_{b,i}}\left| {{e_{\eta 2,i}}} \right|}}{{{B_{2,i}}}}} \right)V_{s2,i}^2 + {e_{\eta 2,i}}{d_{2,i}}} \right)} \\ &\le -\sum\limits_{i = 1}^n {(\frac{{{C_{a,i}}}}{{{B_{2,i}}}}V_{s2,i}^2 - {d_{1,i}})} \left| {{e_{\eta 1,i}}} \right| + {(\frac{{{C_{b,i}}}}{{{B_{3,i}}}}V_{s3,i}^2 - {d_{2,i}})} \left| {{e_{\eta 2,i}}} \right|.\label{51}
\end{aligned}
\end{equation}\end{small}\normalsize
Based on the last line of \eqref{51}, as long as $\frac{{{C_{a,i}}}}{{{B_{1,i}}}}V_{s1,i}^2\ge|{d_{1,i}}|$ and $\frac{{{C_{b,i}}}}{{{B_{2,i}}}}V_{s2,i}^2\ge|{d_{2,i}}|$, the proposed sliding mode control is input to state stable. Based on Assumption 2 and the characteristic of shunting model, the maximum value of $V_{s1,i}$, ${d_{1,i}}$ $V_{s2,i}$, and ${d_{2,i}}$ are $B_{1,i}$ and $\psi_1$, $B_{2,i}$, and $\psi_2$, respectively. Thus, the stability of the bioinspired sliding mode torque control can be guaranteed by tuning these parameters. From \eqref{51}, the estimated velocity error is bounded. Since it is known that the filter provides a precise velocity estimate for its actual value, it concludes that the actual velocity error can be also bounded. This, together with the stability from the bioinspired backstepping control, also implies that the formation tracking error is bounded as well. Thus, the overall stability of the formation system proposed can be guaranteed.
\subsection{Adaptive sliding innovation filter}
First, by letting the mass center and origin of the body fixed frame at the same point, giving the time step $\Delta t$ and using the Euler approximation, the mobile robot dynamic model processed by the ASIF is given by\par
\vspace{-2ex}\begin{small}
\begin{equation}
{{\zeta }_{i,k + 1}} = \left[ {\begin{array}{*{20}{c}}
{{{ \upsilon }_{i,k + 1}}}\\
{{{ \omega }_{i,k + 1}}}
\end{array}} \right] = \left[ {\begin{array}{*{20}{c}}
{{{ \upsilon }_{i,k}}}\\
{{{ \omega }_{i,k}}}
\end{array}} \right] + {{\bar M}_i}^{ - 1}{{\bar B}_i}{\tau _{c,i,k}}\Delta t + {\alpha _{i,k}}
\end{equation}
\begin{equation}
   {{\hat \zeta }_{k + 1}} = {H_i}\zeta _{i,k + 1}  + {\beta _{i,k}}
\end{equation}
\end{small}\normalsize
where ${\alpha _{i,k}}$ and ${\beta _{i,k}}$ are, respectively, the system and measurement noises at time step $k$ of $i$-th robot; ${\upsilon _{i,k + 1} }$ and ${\omega _{i,k + 1} }$ are the linear and angular velocities, respectively; ${{\hat \zeta }_{k + 1}}$ is the state measurements; and $H$ is the measurement matrix. Then, the design of the ASIF shares a similar structure as the KF, which uses a predictor-corrector estimation method. The design procedures of ASIF follow a predicting stage as\par
\begin{small}
\vspace{-3pt}
\begin{equation}
    {{\tilde \zeta }_{i,k + 1|k}} = {A_{i,h}}{{\tilde \zeta }_{i,k|k}} + {B_{i,h}}{\tau _{c,i,k}}
\end{equation}\end{small}
\vspace{-3pt}
\begin{small}\begin{equation}
    {P_{i,k + 1|k}} = {A_{i,h}}{P_{i,k|k}}{A_{i,h}}^T + {Q_{i,k}}
\end{equation}\end{small}
\vspace{-3pt}
\begin{small}\begin{equation}
    {{\tilde z}_{i,k + 1|k}} = {{\hat \zeta }_{i,k + 1}} - {H_i}{{\hat \zeta }_{i,k + 1|k}}
\end{equation}\end{small}\normalsize
where ${{\tilde \zeta }_{i,k + 1|k}}$ and ${{\tilde \zeta }_{i,k|k}}$ are respectively the prior and posterior state estimates at time step $k$; ${P_{i,k + 1|k}}$ is the state error covariance; ${{\tilde z}_{i,k + 1|k}}$ is the innovation; ${A_{i,h}}$ is treated as $2\times2$ identity matrix; $B_{i,h}={{\bar M}_i}^{ - 1}{{\bar B}_i}\Delta t$; and ${Q_{i,k}}$ is the system noise covariance. Then, these predicted variables are processed through the update stage as\par
\noindent
\begin{small}\begin{equation}
        {K_{i,k + 1}} = H_{{{{i,k+1}}}}^ + {\rm{diag}}({\rm{sat}}\left( {\frac{{\left| {{{\tilde z}_{i,k + 1|k}}} \right|}}{\rho_i }} \right)
\end{equation}
\vspace{-2ex}
\begin{equation}
    {{\tilde \zeta }_{k + 1|k + 1}} = {{\tilde \zeta }_{k + 1|k}} + {K_{k + 1}}{{\tilde z}_{i,k + 1|k}}
\end{equation}
\vspace{-2ex}\begin{equation}
\begin{aligned}
      {P_{i,k + 1|k + 1}} = \left( {I - {K_{i,k + 1}}{H_i}} \right){P_{i,k + 1|k}}{\left( {I - {K_{i,k + 1}}{H_i}} \right)^T} \\+ {K_{i,k + 1}}{R_{k + 1}}K_{i,k + 1}^T
\end{aligned}
\end{equation}\end{small}\normalsize
where $\rho_i$ is the sliding boundary layer; ${R_{k + 1}}$ is the measurement noise covariance;${K_{i,k + 1}}$ is the sliding innovation gain; ${{\hat \zeta }_{k + 1|k + 1}}$ is the updated state estimate. Then, a sliding innovation filter should be introduced. In the conventional sliding innovation filter design, the sliding boundary $\rho_i$ is manually tuned. However, this filter only offers sub-optimal solutions under normal operating conditions, since it does not consider system noise covariance to update its state estimates. Thus, the ASIF is developed to have more accurate state estimates. The sliding boundary, sliding innovation gain, and innovation covariance ${S_{i,k + 1}}$, are given by\par
\vspace{-1ex}\begin{small}
\begin{equation}
    {K_{k + 1}} = H_i^ + \left| {{{\tilde z}_{i,k + 1|k}}} \right|\rho _{_{i,k + 1}}^{ - 1}
\end{equation}
\begin{equation}
    {S_{i,k + 1}} = {H_i}{P_{i,k + 1|k}}H_i^T + {R_{i,k + 1}}
\end{equation}
\begin{equation}
    {\rho _{i,k + 1}} = {S_{i,k + 1}}{({S_{i,k + 1}} - {R_{i,k + 1}})^{ - 1}}{\rm{diag}}\left| {{{\tilde z}_{i,k + 1|k}}} \right|.
\end{equation}
\end{small}\normalsize
This completes the overall design of ASIF for mobile robots.

\noindent{\bf Remark 4.} The Kalman filter is capable of providing optimal results if the system is well known; however, if there are modeling uncertainties or modeling errors in the estimator, the Kalman filter fails to provide an accurate state estimate. Thus, the ASIF is introduced, it not only provides accurate state estimates under normal working conditions but also offers extra robustness when modeling errors or uncertainties occur.
\begin{figure}[h]
\centerline{\includegraphics[width=0.40\textwidth]{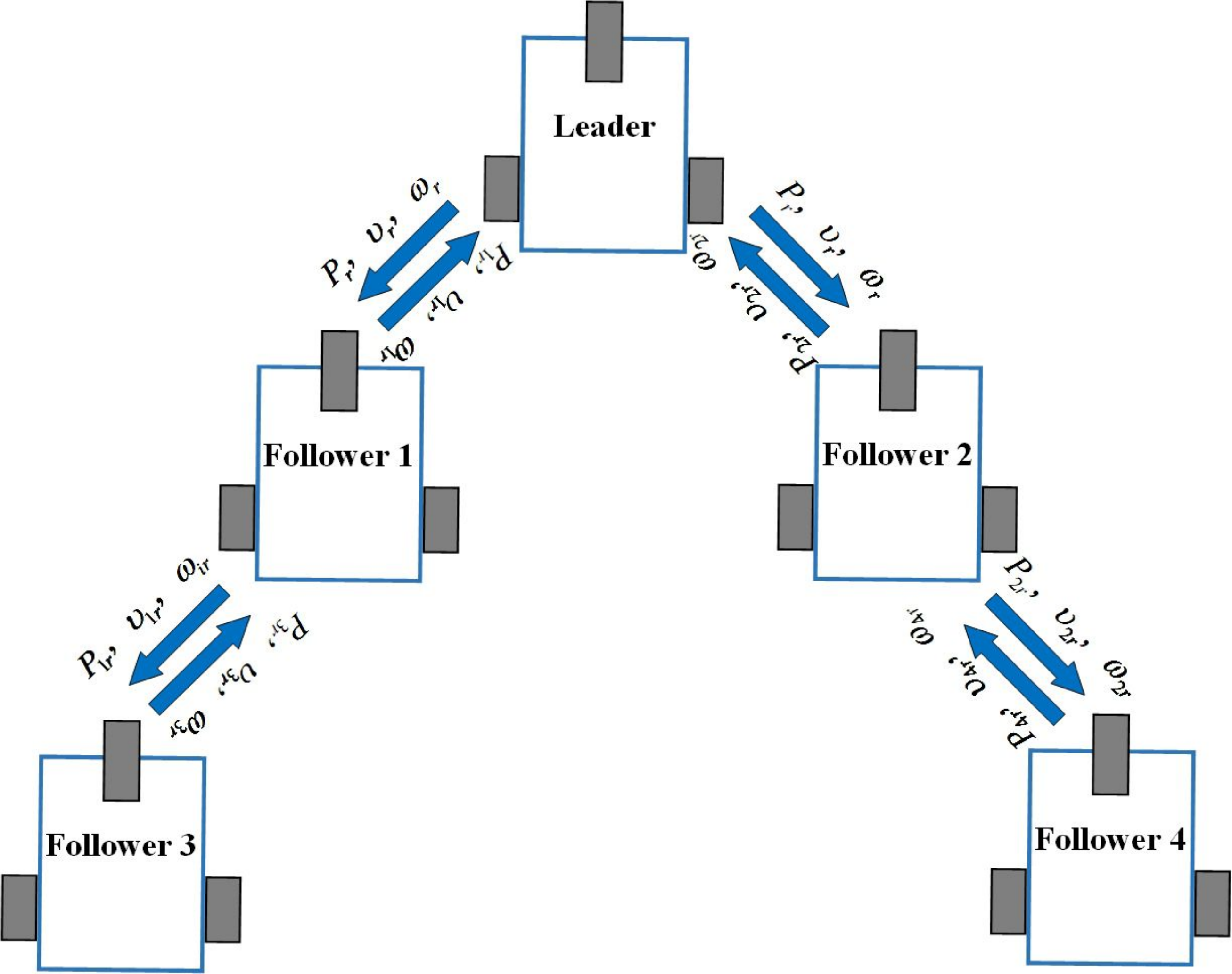}}
\caption{Communication setup and formation of mobile robots}\label{fig2}
\end{figure} 
\section{simulation results}
This section provides various simulation results that demonstrate the control performance of the proposed control method. It is considered that there are four followers and a virtual leader, the communication setups, shown in Figure \ref{fig2}, are assumed to be given. The parameters for the estimator are treated as $k_i=I_{3\times3}$, $k_{a1} = k_{a2} = 20$, $k_{b1} = k_{b2} = 5$. The control parameters in the backstepping controller are treated as $C_{1,i}=3$, $C_{2,i}=2$, and $C_{3,i}=1$, the parameters in the conventional sliding mode dynamic controller are treated as $C_{a,i}=3$ and $C_{b,i}=3$. The parameters of the shunting model in the bioinspired backstepping kinematic controller are treated as $A_i=4$, $B_i=D_i=2$, while the parameters of the shunting model in the bioinspired sliding mode controller are treated as $A_{1,i}=4$, $B_{1,i}=6$, $A_{2,i}=4$, and $B_{2,i}=6$. 

To test the tracking performance of the proposed method, the desired trajectory for the leader to track, is given by ${x_{d}} = t$ and ${y_{d}} = 2 + \sin (\frac{\pi }{2} + t)$. The desired velocities of the leader are calculated by\par
\vspace{-4pt}\begin{small}
\begin{equation}
    {\upsilon _{d}} = \sqrt {{\dot x_{d}}^2 + {\dot y_{d}}^2}\qquad    {\omega _d} = \frac{{{{\ddot y}_d}{{\dot x}_d} - {{\ddot x}_d}{{\dot y}_d}}}{{ {{{\dot x}_d}^2 + {{\dot y}_d}^2} }}
\end{equation}
\end{small}\normalsize
In addition, at the initial stage, the linear velocity follows ${\upsilon _{r}}{\rm{  =  }}{\upsilon _{r}}(1 - {e^{ - t/0.5}})$. The relative positions of the mobile robot are treated as $\Delta x_1 = 4$, $\Delta y_1 = -4$, $\Delta x_2 = 4$, $\Delta y_2 = 4$, $\Delta x_3 = 7$, $\Delta y_3 = -7$, $\Delta x_4 = 7$, $\Delta y_4 = 7$.
\begin{figure}[h]
\centerline{\includegraphics[width=0.45\textwidth]{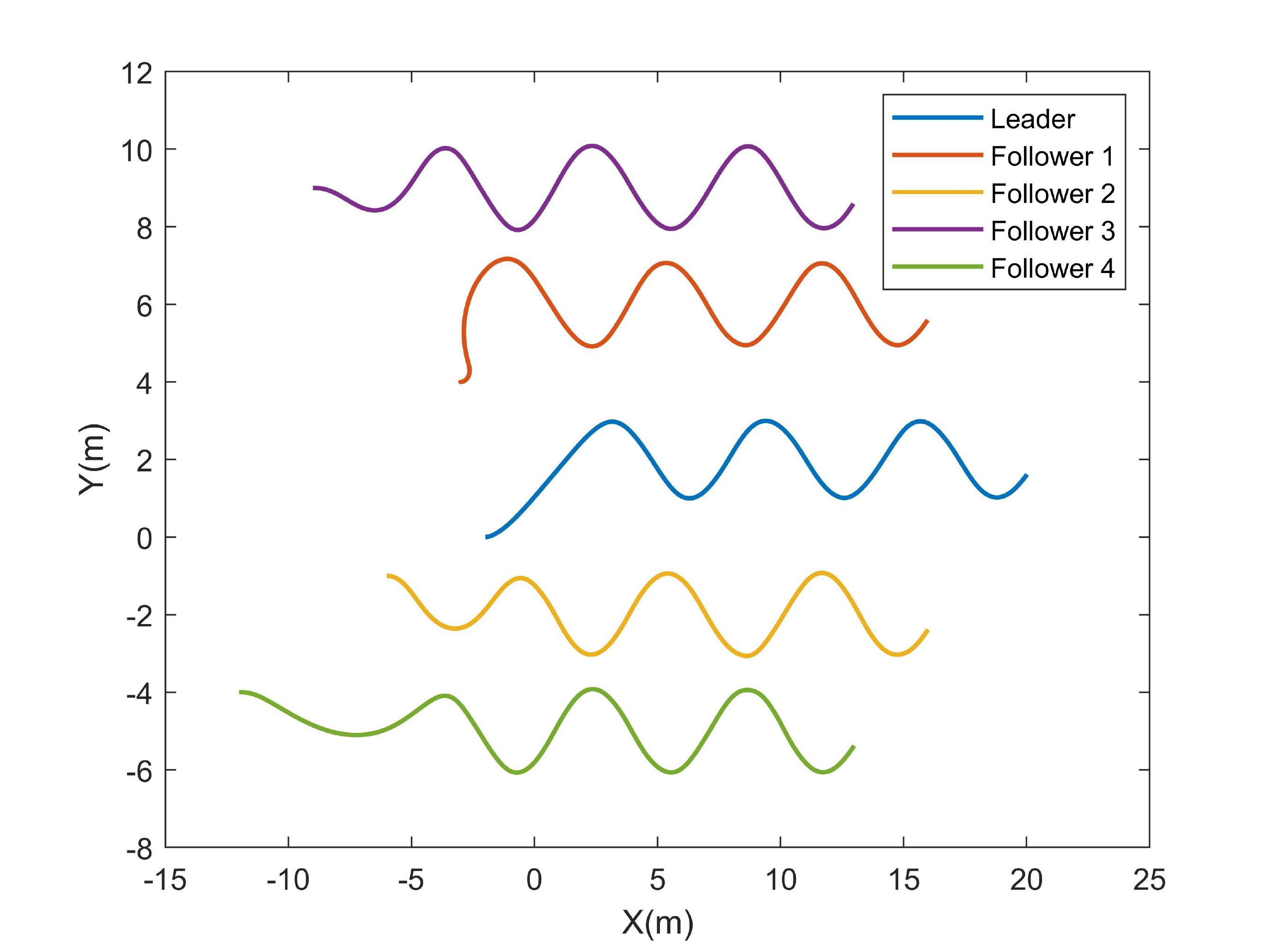}}
\caption{Position of mobile robots tracking a continues path\label{fig3}}
\end{figure} 

\begin{figure}[h]
\centerline{\includegraphics[width=0.45\textwidth]{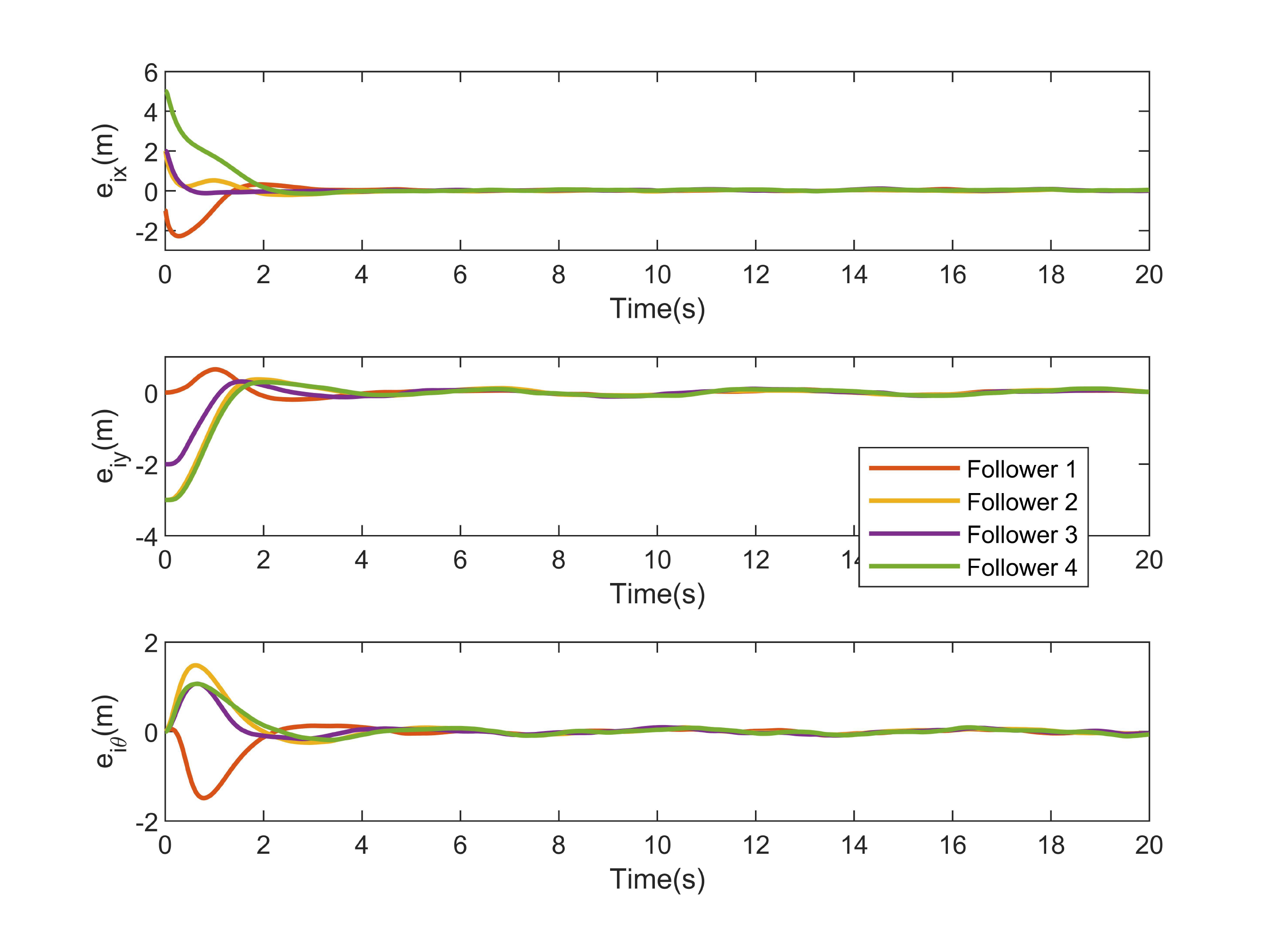}}
\caption{Formation tracking errors of each follower for tracking a continuous path\label{fig4}}
\end{figure} 
As shown in Figure \ref{fig3}. The proposed formation control strategy makes the mobile robots maintain their prescribed formation. Figure \ref{fig4} further shows that the estimation error converges to zero.
\begin{figure}[h]
\centerline{\includegraphics[width=0.50\textwidth]{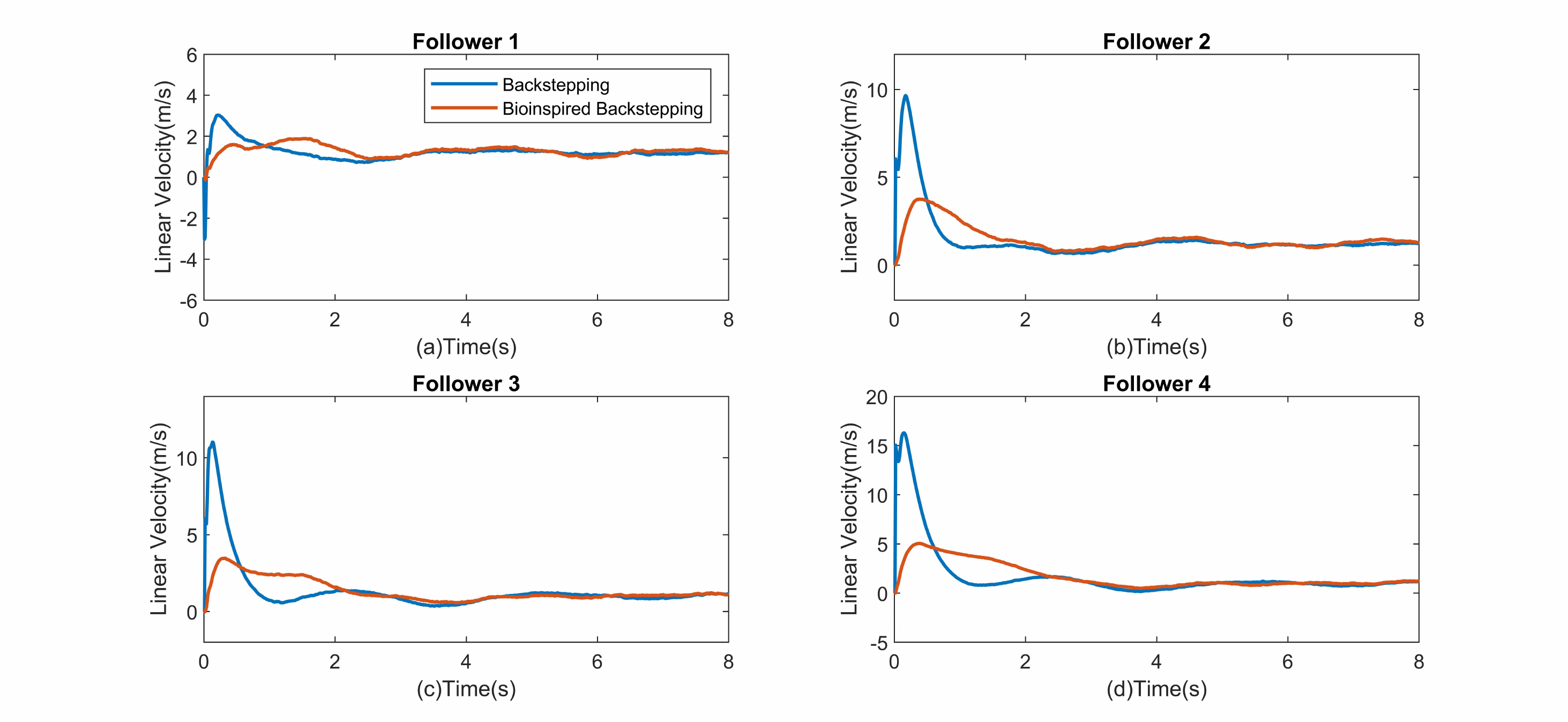}}
\caption{Linear velocity estimates of different methods. Blue: Conventional backstepping \& Bioinspired sliding mode \& ASIF Red: Bioinspired backstepping \& Bioinspired sliding mode \& ASIF }\label{fig5}
\end{figure} 

Furthermore, in Figure \ref{fig5}, when initial tracking errors occur, the conventional backstepping method yields a velocity jump; it is noted that initial tracking errors are calculated as ${\hat x}_{1}=-1$, ${\hat x}_{2}={\hat x}_{3}=2$, and ${\hat x}_{4}=5$, which effectively shows that the larger the initial tracking error, the higher the initial required speed. In addition, the maximum linear velocity that is required by conventional backstepping control reaches 15m/s at its maximum, which is impractical for an autonomous robot operating at that rate of speed. On the other hand, the bioinspired inspired backstepping control has successfully avoided such a linear velocity jump; this implies that an extremely large amount of torque would be needed to drive the mobile robot to reach such a velocity immediately in the conventional backstepping design. Thus, with the help of the proposed bioinspired method, the speed jump issue has been resolved.

Compare to the conventional sliding mode control and second order sliding mode control, the bioinspired sliding mode control provides smooth control inputs under the effects of noise. The unique filtering capability of the shunting model is capable of filtering out high-frequency noise to ease the burden in the control system.
\begin{figure}[t]
\centerline{\includegraphics[width=0.5\textwidth]{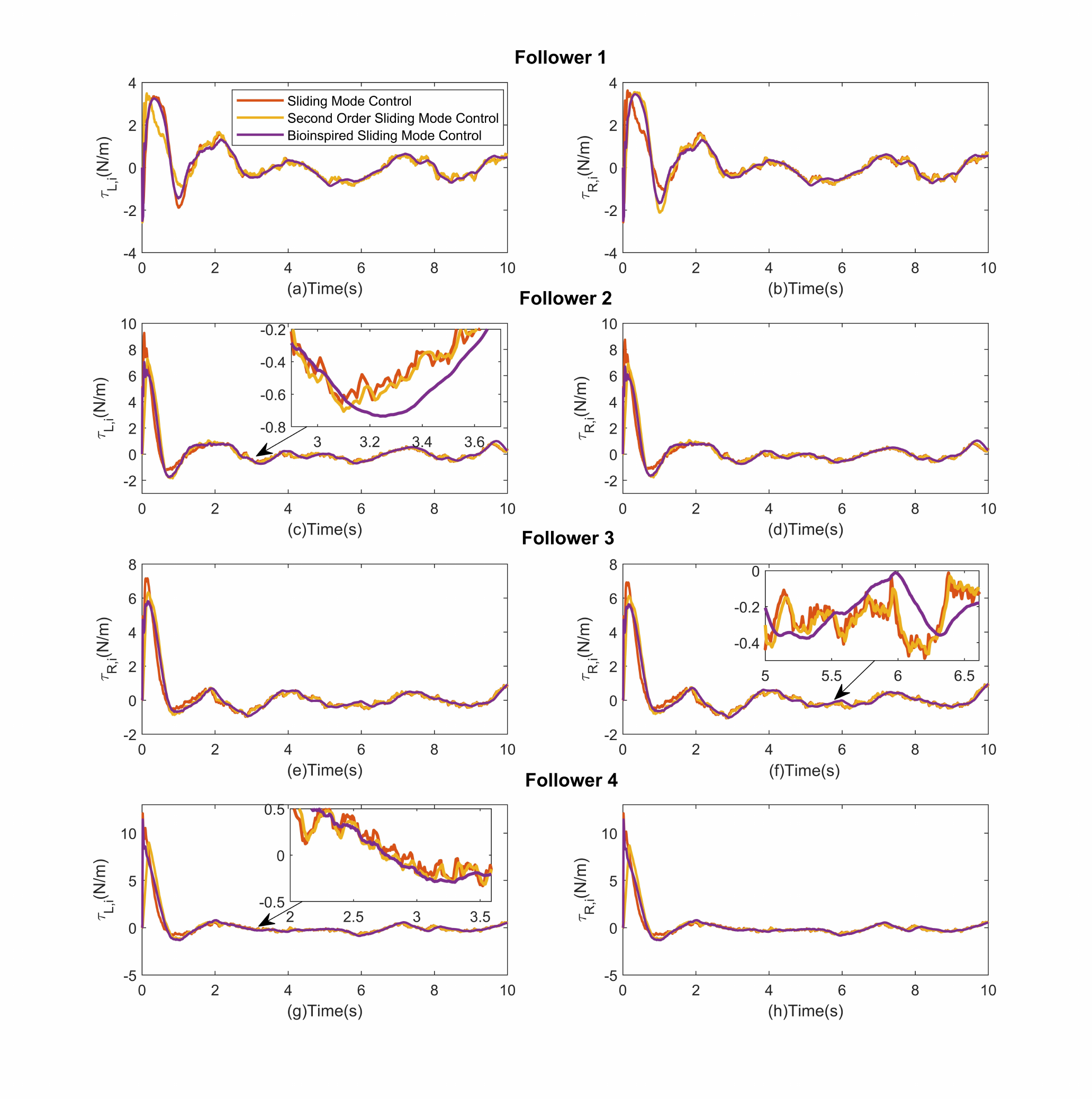}}
\caption{Torque control commands with different control methods. Red: Conventional Sliding Mode Control, Yellow: Second Order Sliding Mode Control, Purple: Bioinspired Sliding Mode Control}\label{fig6}
\end{figure} 

\begin{table}[!t]
\caption{Velocity RMSE under normal case($\times10^{-2}$)\label{tab1}}
\centering
\begin{tabular}{|c |c |c |c|c|c|}
\hline
Linear Velocity &KF &ASIF& Angular Velocity&KF&ASIF\\
\hline
Follower 1&$0.97$ &$0.97$&Follower 1&$0.80$&$0.80$\\ 
Follower 2&$0.96$ &$0.96$&Follower 2&$0.82$&$0.82$\\
Follower 3&$0.97$ &$0.97$&Follower 3&$0.81$&$0.81$\\
Follower 4&$0.98$ &$0.98$&Follower 4&$0.78$&$0.78$\\
\hline
\end{tabular}
\end{table}
\begin{table}[!h]
\caption{Velocity RMSE under faulty case($\times10^{-2}$)\label{tab2}}
\centering
\begin{tabular}{|c |c |c |c|c|c|}
\hline
Linear Velocity &KF &ASIF& Angular Velocity&KF&ASIF\\
\hline
Follower 1&$6.43$ &$1.18$&Follower 1&$2.10$&$1.03$\\ 
Follower 2&$6.50$ &$1.15$&Follower 2&$1.83$&$1.01$\\
Follower 3&$5.37$ &$1.19$&Follower 3&$2.14$&$1.03$\\
Follower 4&$5.75$ &$1.19$&Follower 4&$2.03$&$1.05$\\
\hline
\end{tabular}
\end{table}
The root mean square errors (RMSEs) between the state estimates and the actual states are shown in Table \ref{tab1}. We observe that under normal conditions, the filters work perfectly) and the ASIF provides almost identical state estimates as the Kalman filter does. However, when the condition becomes noisy, the estimator could fail, resulting in inaccurate state estimates. It is assumed that the estimator works perfectly for 10 seconds; after that, there is a fault injected into the system, which makes the mass $m_i$ and inertia $I_i$ that are processed by the filter change to $0.01m_i$ and $I_i=0.1I_i$, respectively. Table \ref{tab2} shows that, in a faulty case, the KF fails to provide accurate state estimates, while the ASIF is still capable of providing relatively more accurate state estimates.

\section{conclusion}
In this paper, the distributed leader follower formation control problem has been addressed. Initially, a distributed estimator is developed to estimate the leader's states in order to maintain a desired formation without needing to know the leader's states for all robots. Then, a distributed bioinspired backstepping control is presented to address the velocity jump problem that makes conventional designs impractical. After that, a bioinspired sliding mode control is proposed to improve the control robustness with smooth torque input. The ASIF is integrated with the proposed control strategy to provide accurate state estimates while maintaining robustness under modeling errors. Finally, the proposed control has been rigorously proven to be stable by Lyapunov theory, and multiple simulation results have demonstrated the effectiveness and efficiency of the proposed formation control strategy.

In the future, a deeper investigation of robot communications such as denial of service attacks and noises that affect the accuracy of the data transmission could be further addressed to improve control effectiveness under various conditions. In addition, more research on obstacle avoidance is expected to improve the overall formation control efficiency.

\bibliographystyle{IEEEtran}
\bibliography{IEEEabrv,references}
\begin{IEEEbiography}[{\includegraphics[width=1in,height=1.25in,clip,keepaspectratio]{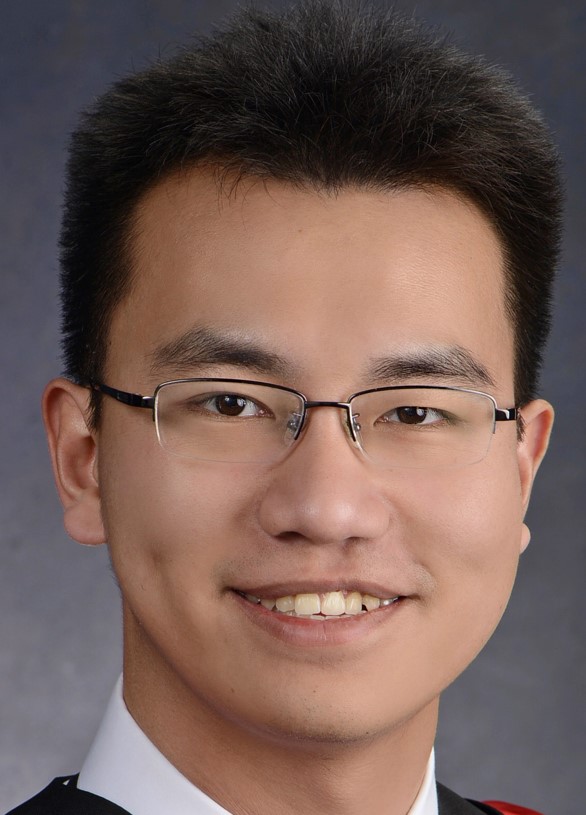}}]{Zhe Xu} (S’20-M'23) received B.ENG. degree in Mechanical Engineering in 2018 and M.A.Sc. degree in Engineering Systems and Computing in 2019 from the University of Guelph. He is currently pursuing the Ph.D. degree in Engineering Systems and Computing with the School of Engineering at the University of Guelph under Professor Simon X. Yang's Supervision. His research interests include networked systems, tracking control, estimation theory, robotics, and intelligent systems.
\end{IEEEbiography}
\begin{IEEEbiography}[{\includegraphics[width=1in,height=1.25in,clip,keepaspectratio]{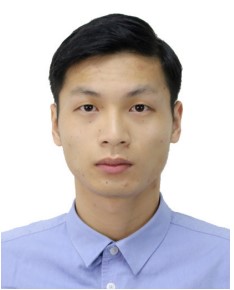}}]{Tao Yan} (S’22) received the B.Sc. degree from
the North China Institute of Aerospace Engineering,
Langfang, China, in 2016; the M.Sc. degree from
the Zhejiang University of Technology, Hangzhou,
China, in 2020. He is currently pursuing his Ph.D.
degree at the University of Guelph, ON, Canada.
His research interests include the intelligent control,
distributed control and optimization, and networked
underwater vessel systems.
\end{IEEEbiography}
\begin{IEEEbiography}[{\includegraphics[width=1in,height=1.25in,clip,keepaspectratio]{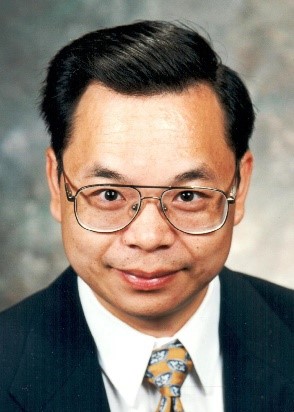}}]{Simon X. Yang} (S'97--M'99--SM'08) received the B.Sc. degree in engineering physics from Beijing University, Beijing, China, in 1987, the first of two M.Sc. degrees in biophysics from the Chinese Academy of Sciences, Beijing, in 1990, the second M.Sc. degree in electrical engineering from the University of Houston, Houston, TX, in 1996, and the Ph.D. degree in electrical and computer engineering from the University of Alberta, Edmonton, AB, Canada, in 1999.  Dr. Yang is currently a Professor and the Head of the Advanced Robotics and Intelligent Systems Laboratory at the University of Guelph. His research interests include robotics, intelligent systems, sensors and multi-sensor fusion, wireless sensor networks, control systems, machine learning, fuzzy systems, and computational neuroscience. 

Prof. Yang has been very active in professional activities. He serves as the Editor-in-Chief of International Journal of Robotics and Automation, and an Associate Editor of IEEE Transactions on Cybernetics, IEEE Transactions of Artificial Intelligence. 
\end{IEEEbiography}
\begin{IEEEbiography}[{\includegraphics[width=1in,height=1.25in, clip,keepaspectratio]{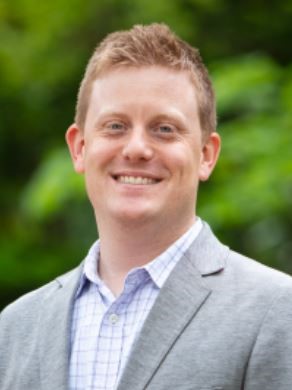}}]{S. Andrew Gadsden} (M'09--SM'19) is an Associate Professor in the Department of Mechanical Engineering at McMaster University. His research area includes control and estimation theory, artificial intelligence and machine learning, and cognitive systems. He earned his PhD in Mechanical Engineering at McMaster University. Andrew was an Associate/Assistant Professor at the University of Guelph and the University of Maryland. Andrew is an elected Fellow of ASME, is a Senior Member of IEEE. He is also a certified Project Management Professional. Andrew is an Associate Editor of Expert Systems with Applications and is a reviewer for a number of ASME and IEEE journals and international conferences.

Andrew has been the recipient of numerous international awards and recognitions. In January 2022, Andrew and his fellow air-LUSI project teammates were awarded NASA’s prestigious 2021 Robert H. Goddard Award in Science for their work on developing an airborne lunar spectral irradiance

\end{IEEEbiography}
\end{document}